# Efficient Induction of Finite State Automata

Matthew S. Collins and Jonathan J. Oliver
Computer Science Dept,
Monash University
Clayton, Victoria 3168, Australia

## Abstract

This paper introduces a new algorithm for the induction of complex finite state automata from samples of behaviour. The algorithm is based on information theoretic principles. The algorithm reduces the search space by many orders of magnitude over what was previously thought possible. We compare the algorithm with some existing induction techniques for finite state automata and show that the algorithm is much superior in both run time and quality of inductions.

## 1. INTRODUCTION

This paper presents a new search algorithm for the induction of Probabilistic Finite State Automata (PFSA) from positive samples of behaviour. The Information (theory) Guided Search (IGS) algorithm applies Minimum Message Length (MML) based criteria to form a highly effective PFSA induction procedure. The concept of *partial MML* is introduced as a mechanism for partitioning the plausible model space to permit efficient examination. This partitioning of the model space allows the IGS algorithm to refute the previously held belief that "only an exhaustive search of all the possible PFSA for a given data set can ensure the truly optimal PFSA" (via an MML criterion) "has been found." (Patrick and Chong (1987)) The IGS algorithm shows this belief to be incorrect by many orders of magnitude. We present an example that requires 39,541,447 automata to be exhaustively searched, however the IGS algorithm is able to find and prove it has the optimal machine after examining only 85 nodes[1].

In this paper we present the MML foundations that the criteria are based on, however we do not present the criteria. There are a number of criteria the IGS algorithm may use. Examples of these criteria are given by Wallace and Georgeff (1983)[2], Collins and Zukerman (1994), Collins and Oliver (1997a). The IGS algorithm is then presented followed by its comparison with existing PFSA induction techniques. More detail about the IGS algorithm and results may be found in Collins and Oliver (1997b).

## 2. MML BASIS

Given a set of plausible hypotheses about the source of some given body of data (***D***), it is desirable to select the hypothesis (***H***) that is most likely the true explanation of ***D***. In the event that none of the hypotheses is the true explanation then we wish to select the hypothesis that is "closest" to the true explanation. The MML principle is an information theoretic based approach to providing criteria for the comparison of hypotheses about ***D***. The MML principle was first introduced by Wallace and Boulton (1968) and was later more formally expressed by Wallace and Freeman (1987). Oliver and Hand (1994) provide an introductory paper to the topic. The MML principle has many similarities to Bayesian inference methods (Oliver and Baxter (1994)).

For discrete structures[3] to find the most probable ***H*** given ***D*** within the Bayesian framework requires that we maximise:

$$\mathrm{P}(\mathrm{H}|\mathrm{D}). \tag{1}$$

Applying Bayes' rule to equation (1) yields:

$$\mathrm{P}(H|D)=\frac{\mathrm{P}(H)\times \mathrm{P}(D|H)}{\mathrm{P}(D)}. \tag{2}$$

Given ***D*** is constant and hence **P(*D*)** is the same for all hypotheses to be tested it is only necessary to maximise:

$$\mathrm{P}(H)\times \mathrm{P}(D|H). \tag{3}$$

Taking the negative logarithm yields:

$$-\log(\mathrm{P}(H))-\log(\mathrm{P}(D|H)). \tag{4}$$

Equation (4) must now be minimised.

From information theory an optimal encoding scheme encodes a binary message where each element is encoded with $-\log_2(\mathrm{P}(\text{element}))$ bits. Special techniques exist to ensure that the entire message length is within one bit of the theoretical optimum.

The message length of an optimally encoded message containing a hypothesis followed by an encoding of the data given that hypothesis is the same as given by equation (4) (using base 2 logs). MML ranks hypotheses in the same way as the Bayesian ranking given by equation (2). The most probable hypothesis has the shortest message length. This is the fundamental basis of *Minimum Message Length* (MML) encoding as a method for inferring models from data.

[1] Each node is approximately equivalent to examining one automaton.

[2] Discussed in detail in Collins and Oliver (1996)

[3] The MML paradigm is distinct from the standard Bayesian paradigm when estimating real valued parameters. See Wallace and Freeman (1987) for details. For the induction of automata we are primarily concerned with the inference of discrete properties.

# 3. INFERENCE TECHNIQUES FOR AUTOMATA

There are many techniques for the inference of finite state automata. Many of these, such as Gold (1972), Angluin (1987), Hachtel *et al.* (1991), Schapire (1992), Kam *et al.* (1994) or Oliveira and Edwards (1995) require both negative and positive examples or input output responses of the automaton being examined. Some also require the ability to ask membership questions. These algorithms are state minimisation algorithms that attempt to find the automaton of minimum number of states that is consistent with the observed data. As this is directly implied by the data and there is only one minimum consistent automaton no induction takes place, this is strictly a search problem. These algorithms do have very practical applications for the simplification of state machine design but are not suited to inferring grammars from positive data (or just strings of observed output from the process being examined). The algorithms that are capable of this type of inference can be grouped into two main approaches.

## 3.1 CRITERION BASED APPROACHES:

The criterion based approach seems to have first been successfully applied by Gaines (1976). Gaines (1976) defined two measures for an automaton; a measure of complexity (the number of states) and a measure of degree of fit (negative log likelihood that the machine generated the data). All possible machines are searched and the degree of fit and complexity are traded off to generate an admissible set of likely machines. An automaton is judged admissible if it has a better degree of fit than all other machines with the same or simpler complexity. Gaines was unable to formally compare between models in the admissible set and argues for the machine with the machine with be largest increase in degree of fit for a small increase in complexity. Witten (1980) modifies the technique of Gaines by essentially altering the measure of complexity to be the number of arcs in the machine and therefore still has the problem that he is unable to formally distinguish between members of the admissible set.

Maryanski and Booth (1977) provide another criterion for selecting a particular automaton. Their algorithm enumerates the possible automata in order of simplicity (minimum number of arcs) and performs a chi-squared test over the observed data and expected distribution of strings from the trial automata. The first automaton that passes this test is accepted. The user must select the threshold at which the chi-squared test is to pass. The threshold defines the balance between degree of fit and model complexity. From all values of the threshold an admissible set could be constructed for which there is no way to formally argue which is preferred.

The Minimum Message Length based criteria of Wallace and Georgeff (1983) solves the problem of how to trade model complexity and degree of fit. As with the other criteria based approaches Wallace and Georgeff (1983) are still required to search over all possible automata to find the most likely automaton. The IGS algorithm avoids the need for exhaustive search.

## 3.2 EQUIVALENCE BASED APPROACHES

Equivalence based approaches draw their inspiration from the Nerode (1958) realisation technique for synthesising FSA and are based on the concept of defining an equivalence relation (or function) that is used to judge if two states should be merged or not. A classic example of this is the $k$-tails algorithm of Biermann and Feldman (1972) where two states are merged if the set of strings up to length $k$ producible from each state exactly agree.

The usual method for applying the equivalence based approaches are;

1) define an equivalence relation,
2) construct a canonical automata,
3) apply the equivalence relation to reduce machine,
4) repeat step 3 as necessary.

The canonical automata is a tree representation of the data and makes the minimal amount of assumptions about the data.

The algorithms that follow this approach essentially differ in the equivalence relation suggested. Examples are the $k$-tails algorithm (Biermann and Feldman (1972)), Length-$k$ models (Witten (1979)), the *tail-clustering* algorithm (Miclet (1980)), the *successor method* (Vernadet *et al* (1982)), the *predecessor and successor method* Richetin and Vernadet (1984) and a generalised *predecessor and successor method* (Kudo and Shimbo (1988)).

The problem with these approaches is that there is no formal argument given as to why applying an equivalence relation is appropriate for automaton induction. This explains the proliferation of the different techniques. All the techniques require the user to specify some parameters (for example the length of $k$) and generally produce wildly different results for different parameter settings. The user has no way to differentiate between the machines produced for different settings and simple chooses one that looks good.

There are two exceptions that fall somewhere between the two approaches and these are given by Patrick and Chong (1987) and *sk*-strings algorithm of Raman and Patrick (1995). The algorithm of Patrick and Chong (1987) represents a tight coupling of the two approaches where each merge as selected by a set of heuristics is tested using the MML criterion of Wallace and Georgeff (1983). If the MML is not improved the merge is rejected and another merge tried.

The method of Raman and Patrick (1995) appears to supersede the method of Patrick and Chong (1987) and represents a loose coupling. The problem with the Patrick and Chong (1987) technique is summarised by Raman and Patrick (1995) who have concluded that "the method is not likely to always succeed for it is known that in several cases a *good* overall merge might consist of several *bad* component merges." The *sk*-strings algorithm appears to be a hybrid of the *k*-tails algorithm and the *tail-clustering* algorithm of Miclet (1980). In addition the *sk*-strings algorithm defines a number of equivalence relations, a result for each one is generated and the best selected using an MML criterion. The number of candidate automata that are tested with the MML criterion is limited (about 6), hence the algorithm is dominated by the problems inherent with the standard equivalence based algorithms.

Note that most of the algorithm discussed in this section are discussed in more detail in Collins and Oliver (1997b).

# 4. IGS ALGORITHM

## 4.1 PFSA GENERATION

Consider the body of data: *D*="CAAAB/BBAAB/CAAB/ BBAB/CAB/BBB/CB/". There are 39,541,447 possible deterministic PFSA where all arcs have one or more transitions that could be a possible explanation of *D*. Most researchers generally reach the same conclusions about the grammar that was likely to have induced the data set with some minor exclusions. The general consensus is that the PFSA in Fig. 1 is the most plausible explanation of the data.

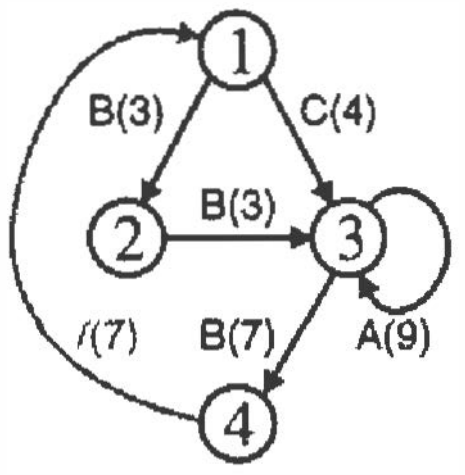

Fig. 1: Optimal PFSA.
*D*="CAAAB/BBAAB/CAAB/BBAB/CAB/BBB/CB/".

To apply a criterion based approach it is necessary to enumerate all possible PFSA. There are several techniques for the enumeration of all the possible PFSA. The approach presented here is a new approach, however it has some similarities to the approach of Gaines (1976).

Consider all the strings of data in parallel:

```
C A A A B /
B B A A B /
C A A B /
B B A B /
C A B /
B B B /
C B /
```

Fig. 2: Parallel analysis of first element in each string.

From the data it can be observed that all deterministic machines in the set of possible PFSA must have at least a 'C' and a 'B' arc originating from the first state. The 'C' arc must have at least 4 transitions and the 'B' arc 3 transitions. Further more, it is known from which strings these transitions come from. A *construction tree* is built where the root node contains all this information. Fig. 3 shows the root node of the construction tree.

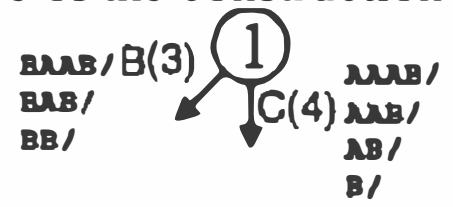

Fig. 3: Root node of construction tree.

If the 'C' arc originating from the first state is now considered then all PFSA in the set of possible PFSA can be divided into two *families*; the family where that arc returns to the first state and the family where it goes to another state. The tree is partitioned as shown in Fig. 4.

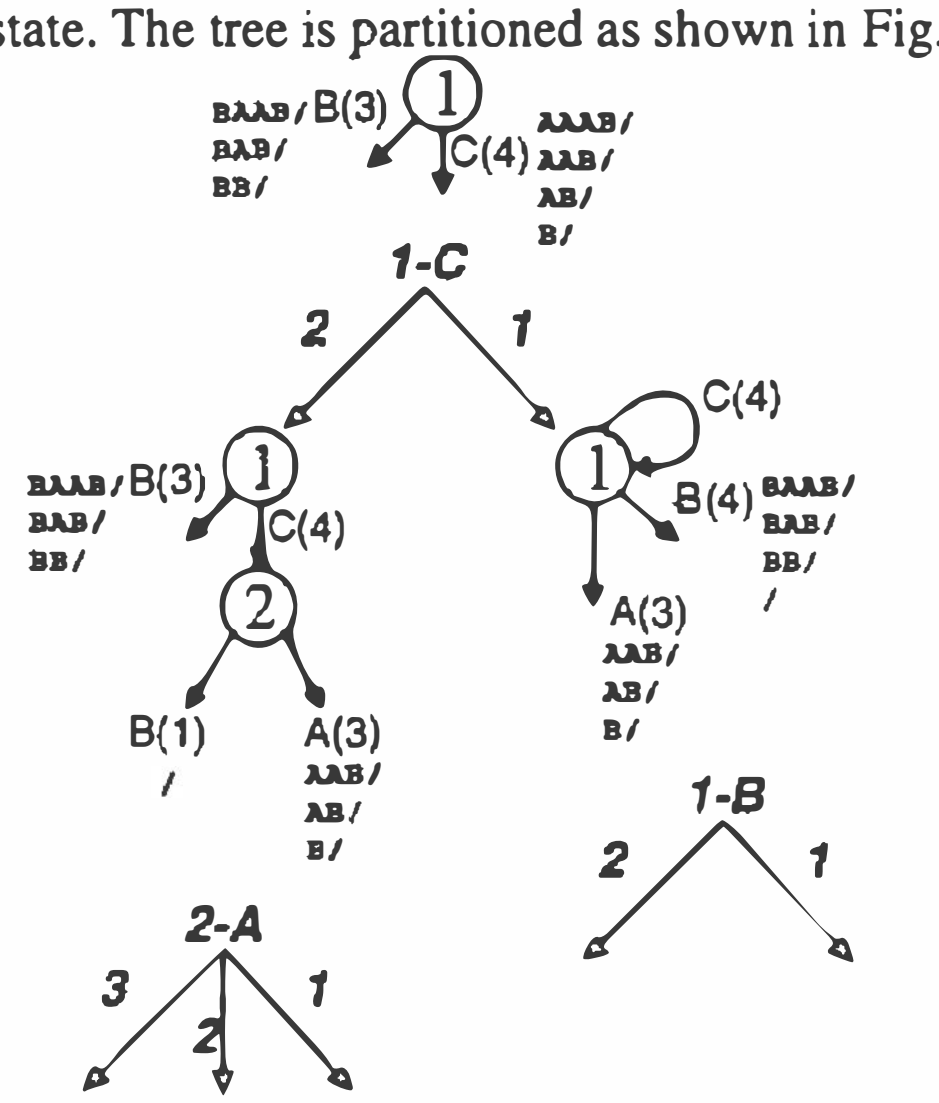

Fig. 4: Tree partitioning.

This process of defining automata sub families by partitioning continues until there are no more *dangling arcs*. A dangling arc is an arc that has no defined destination state but known traversals of data. The entire tree for the given data has 44,199,227 nodes. There are 39,541,447 complete PFSA in the tree (the leaves of the tree) and 4,657,780 *partial PFSA*.

## 4.2 PROBABLE SEARCH SPACE

For a brute force search, the MML of all the leaves of the construction tree are compared. The construction tree however provides a lot of structural information about the families of PFSA. We consider the problem of given a complete PFSA, how many PFSA in the construction tree must be examined to prove the MML of the given PFSA is the lowest possible?

Ignoring the data remaining to be encoded for a partial PFSA in the construction tree the following is true about the family of complete PFSA descendant from a partial PFSA:

1) All family members have the same or more states.

2) All family members have the same or more arcs on each state.

3) All family members have the same or more transitions of data on each arc.

We now take advantage of a property of the MML measure that does not appear to have been previously used. The MML measure is a monotonically increasing measure with complexity. That is, the following can be said about the MML measure for PFSA:

1) Adding additional transitions and keeping the structure the same will increase the MML.

2) Adding additional structural elements to the PFSA and keeping the number of transitions on the existing arcs the same will increase the MML.

A *partial MML* is computed by evaluating the MML measure to the extent permitted by the partial PFSA. It is known that the true MML of all families descendant from a node must be greater than the partial MML. If the partial MML is greater than the current best MML then the node and all families descendant can be safely discarded.

The dramatic effect of the reduction of search space this method of culling has can be seen by example. Using the data given in Fig. 3 and starting with the 1-state machine as an initial hypothesis and using a breadth first search, 269 nodes were required to be examined to prove the PFSA in Fig. 1 was optimal. We define this area of nodes in the construction tree as the *probable search* space. It can be seen that for this example the probable search space is already more than 5 orders of magnitude smaller than the possible search space.

## 4.3 PROBABLE SEARCH SPACE PROPERTIES

A number of factors effect the size of the probable search space.

1) Measure Efficiency: The criteria's efficiency of separating PFSA in part determines the size of the probable search space. The efficiency of the measure, MML or otherwise[4] is determined by how few nodes need to be expanded for partial MML of poor quality PFSA to reach the cost of the final MML of good quality PFSA. The current best criteria are given by Collins and Oliver (1997a).

2) Arc Expansion Order: After selecting a partial PFSA to expand there may be a number of dangling arcs that could be selected next. To maximise the rate at which nodes are deleted from the construction tree the rate that the partial MML approaches the real MML must be maximised. At present we select the dangling arc with most transitions.

3) Node Selection Order: To reduce the amount of work done searching the construction tree the rate at which better solutions are located is critical. The next section discusses heuristics for searching the construction tree. For the given example using these heuristics it is possible to reduce the probable search space to its minimum of 85 nodes.

## 4.4 HEURISTIC DRIVEN SEARCH

The heuristics for selecting the next node to expand are critical to the performance of the search algorithm when the current best PFSA is sub-optimal.

### 4.4.1 MML Estimate Heuristic

The simplest heuristic driven approach is to estimate the best final MML for each node then to expand the node with the lowest estimated MML. Fig. 5 shows the relationship between the fraction of data encoded and the partial MML as the nodes of the construction tree are descended to reach a final complete PFSA. The final PFSA given in the example was a 64 state machine. The strongly linear relationship between the fraction of data consumed and the partial MML has been observed in all the test cases.

The approach that has been taken that yields better results than linear extrapolation is to compute a graph as shown in Fig. 5 for the current best PFSA and to estimate the final MML of a node as the partial MML plus the difference in the fraction remaining as given by the graph.

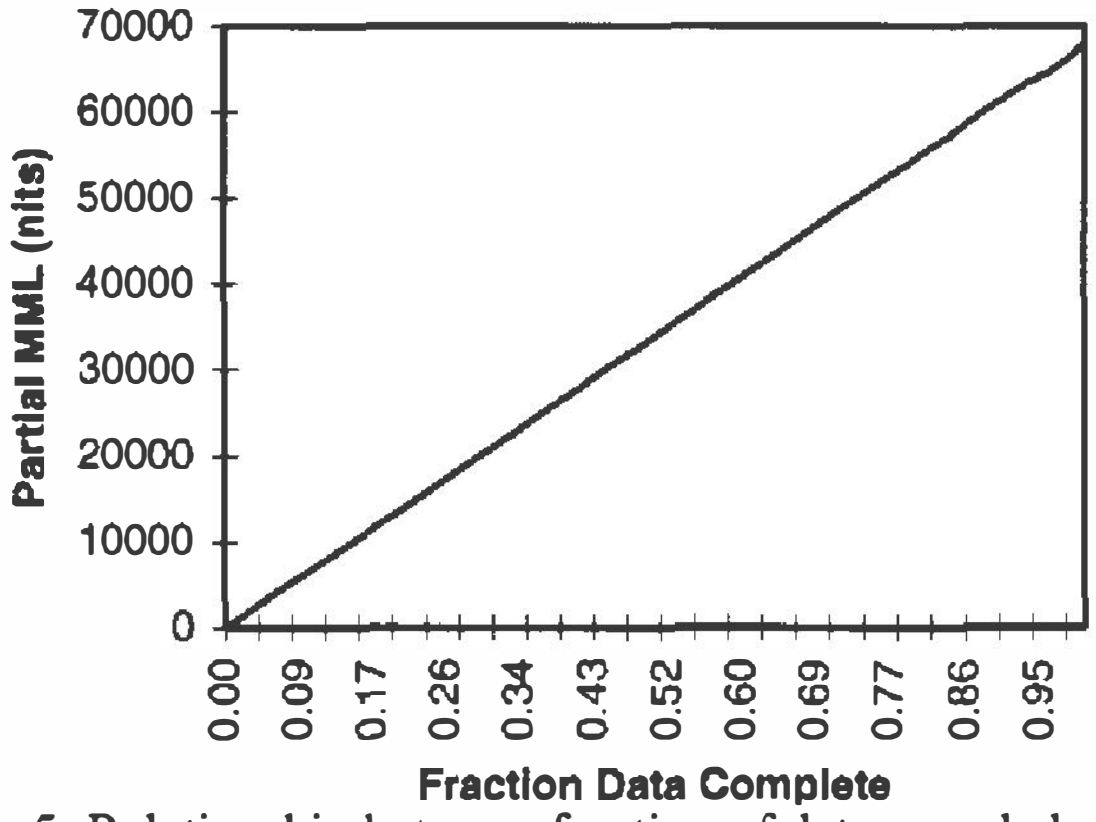

Fig. 5: Relationship between fraction of data encoded and partial MML[5].

Using this approach the 12 state PFSA as shown in Fig. 6 was able to be inferred by examining only 293 nodes in the construction tree. The machine is identical to the original machine used to generate the data.

A problem that is often encountered with this heuristic is that it fails to complete PFSA that would have an MML lower that the current best MML. The heuristic often prefers to expand other incomplete PFSA.

We consider a more difficult problem to those that have been considered so far. The data is drawn from

4 Any measure is applicable to this algorithm on the condition that it can be evaluated for partial PFSA and that it is monotonically increasing.

5 Note: 1 nit = 1/(loge(2)) bits.

classifications of protein secondary structure. The data set consists of 75 secondary structure strings of various proteins. Table 1 is an example of a such a string.

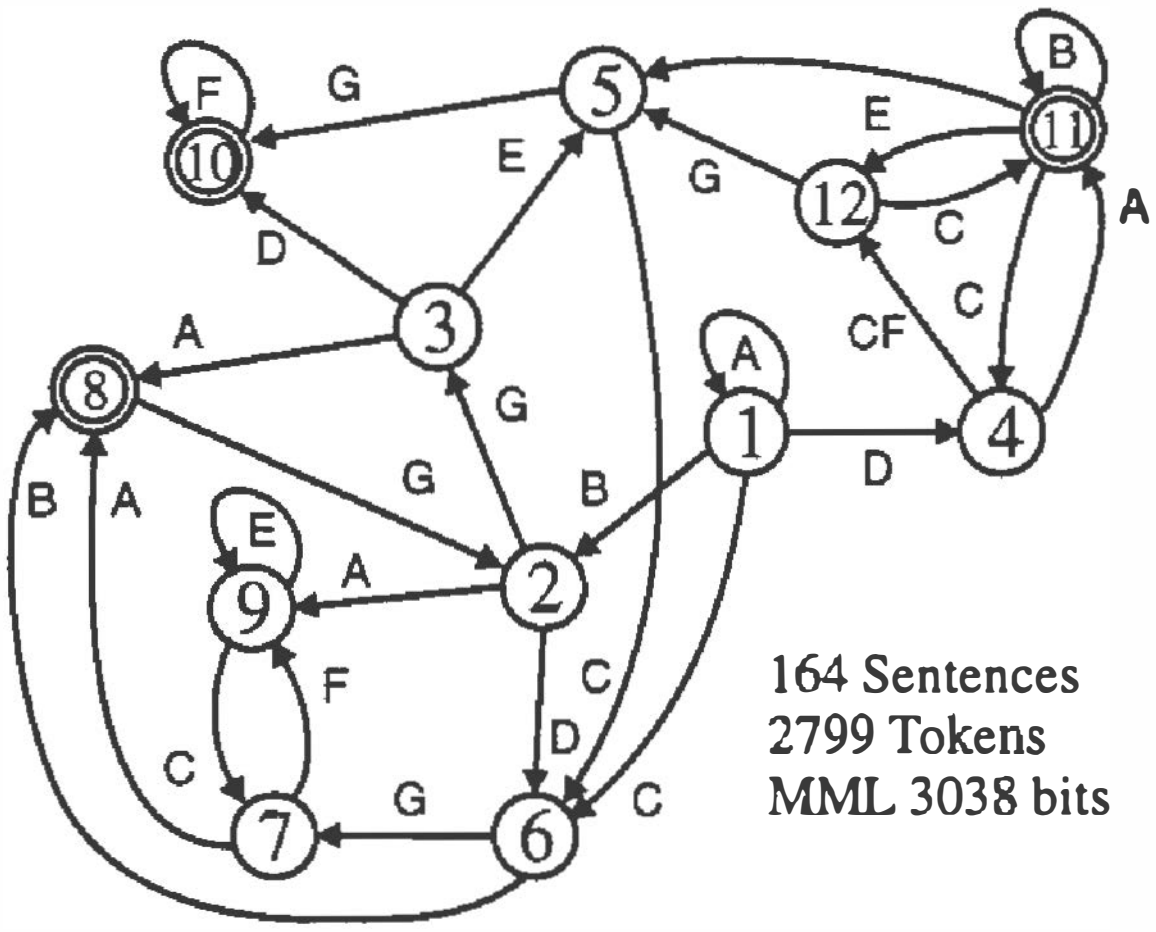

Fig. 6: 12 State inferred PFSA. Double circles represent delimited states.

Table 1: Sample protein secondary structure string.

```
%0=Other,1=Extended,2=Helix,3=Turn,4=delimiter
%protasea
0 0 0 3 0 1 1 1 1 3 3 1 1 1 1 0 0 1 1 1 1 1 3 3
1 1 1 1 1 1 0 2 2 2 2 3 3 0 0 1 1 3 3 1 1 1 1 1
1 1 0 0 0 0 0 1 1 1 1 1 1 0 0 2 2 2 0 0 0 1 1 1
0 0 0 0 0 1 1 1 0 0 1 1 0 0 0 0 3 3 0 1 1 1 1 1
1 3 3 3 1 1 1 1 1 1 1 1 1 1 1 1 1 1 1 0 2 2 2 0
1 1 1 1 1 1 1 1 0 0 0 0 0 3 3 0 3 3 0 1 1 1 1 3
3 1 1 1 1 1 1 1 1 1 1 1 1 3 3 3 1 1 1 1 1 1 1 1
2 2 2 2 2 2 2 2 3 1 1 1 0 4
```

When the program is run with the current heuristic running on a DEC Alpha the best result achieve in a 1000 second period is shown in Table 2.

Table 2: Secondary Structure Test Result: MML Estimate heuristic.

```
There are 8 states with a max of 5 arcs
Automata cost is: 13833.10408bits
arc->   O    E    H    T   d
 state
   0   [1]   -    -    -   -
   1   [1]  [2]  [4]  [1] [0]
   2    -   [3]   -    -   -
   3   [1]  [3]  [4]  [7]  -
   4    -    -   [5]   -   -
   5    -    -   [6]   -   -
   6   [1]  [2]  [6]  [1]  -
   7   [2]   -    -   [1]  -
Nodes examined 19430, Completed PFSA 15
CPU (200Mhz Alpha) time: 0:00:16
```

#### 4.4.2 Compression Heuristic

The compression heuristic is used to encourage completion. The heuristic selects the node to expand that can compress the remaining data the least efficiently to have a final message length the same as the current best. Let $Z$ be the number of remaining tokens unencoded for a given node in the construction tree. The value of the compression heuristic is given as:

$$\frac{(\text{partial MML} - \text{current best MML})}{Z}.$$

The node with the smallest value is selected. The program was run using the compression heuristic on a 486-66 with a 1000 second time out. The result is shown in Table 3.

Table 3: Secondary Structure Test Result: Compression heuristic.

```
There are 13 states with a max of 5 arcs
Automata cost is: 12925.37846bits
arc->   O    E    H     T    d
 state
   0   [1]   -    -     -    -
   1   [1]  [2]  [5]   [8]  [0]
   2    -   [3]   -     -    -
   3   [1]  [4]  [10]  [11]  -
   4   [1]  [7]  [10]  [11]  -
   5    -    -   [6]    -    -
   6   [1]  [2]  [6]   [8]   -
   7   [1]  [9]  [10]  [11]  -
   8   [1]  [2]  [10]  [1]   -
   9   [1]  [3]  [10]  [12]  -
  10    -    -   [5]    -    -
  11   [9]  [2]  [10]  [4]   -
  12    -    -    -    [11]  -
Nodes examined 7831, Completed PFSA 734
CPU (486-66) time: 0:00:11
```

When comparing the results from Table 2 and Table 3 it can be seen that the MML of the PFSA in Table 3 (12925 bits) is lower than that given in Table 2 (13833 bits). It can also be seen that the PFSA in Table 2 is much cleaner than the PFSA in Table 3. The reason the MML is lower for the PFSA in Table 2 is that some structure with the turns is starting to be discovered. It can be seen from the example that about 9% of all nodes examined with the compression heuristic are complete PFSA whereas less than 0.08% of the nodes examined for the MML estimate heuristic were complete PFSA.

### 4.5 SWITCHED HEURISTICS

Whilst the compression heuristic gave a better result there were probably many partial PFSA in the construction tree that could have been quickly finished off to yield a lower MML with the MML estimate heuristic. To allow the heuristics to complement each other they are run in cycles. The MML estimate heuristic is run for about 200 node expansions then the compression heuristic is used for the next 67 expansions and so on.

Table 4 shows some results obtained by switching the two heuristics. The time-out period was 1000 seconds on a 486-66.

The switched heuristics became less trapped in local minima and were able to continue finding new PFSA for longer than either heuristic used independently.

Table 4: Secondary Structure Test Result: Switched heuristics 3:1.

```
There are 9 states with a max of 5 arcs
Automata cost is: 12756.31962bits
arc->    O     E     H     T    d
 state
   0    [1]    -     -     -    -
   1    [1]   [2]   [5]   [4]  [0]
   2     -    [3]    -     -    -
   3    [1]   [3]   [5]   [7]   -
   4    [1]   [2]   [5]   [1]   -
   5     -     -    [6]    -    -
   6    [1]   [2]   [6]   [4]   -
   7    [8]    -     -    [8]   -
   8    [1]   [2]   [5]   [8]   -
Nodes examined 87239, Completed PFSA 5670
CPU (486-66) time: 0:04:26
```

## 4.6 COMPATIBILITY TESTING.

When expanding a node, it is possible however to decide that some of the destination states for an arc are unlikely to yield good PFSA. The child nodes that are unlikely can be immediately discarded without ever having to compute the MML.

The *compatibility test* is performed by comparing the distribution of data from the dangling arc and the distribution of transitions from each of the possible destination states. If the two distributions are similar enough then the node is permitted otherwise it is omitted from the set of children. To test if two distributions are compatible an MML test is used[6]. The message length of the sum of the distributions is computed and if this is greater than the sum of the message lengths of the two distributions then the node is rejected.

For a given distribution let there be ***A*** possible classes, ***a*** of which are present. It is assumed that ***A*** is know *a priori*. We assume that the total number of things (***t***) to be encoded is also known *a priori*. To state the value of ***a*** and to select which of the ***A*** classes these are has the cost given be equation (5).

$$\log_e(A)+\log_e\left(\frac{A!}{(A-a)!\times a!}\right)\text{nits.} \qquad (5)$$

For each class let $n_i$ be the number of occurrences of each class member. The cost required to state the data is given by equation (6).

$$\log_e\left(\frac{(t+a-1)!}{(a-1)!}\right)-\sum_{i=1}^{i=A}\log_e(n_i!)\ \text{nits.} \qquad (6)$$

Note that from Boulton and Wallace (1969) equation (6) has approximately the same message length as if it had been encoded using an MML message.

6 Minor errors in the test are noted in Collins and Oliver (1997b), these errors however tend to cancel and the measure is a good approximation.

Table 5 shows the result using the compatibility testing on the secondary structure data. The number of node examined includes the number of nodes discarded by the compatibility test.

Table 5: Secondary Structure Result: Switched heuristics 3:1, compatibility testing.

```
There are 10 states with a max of 5 arcs
Automata cost is: 12678.68247bits
arc->    O     E     H     T    d
 state
   0    [1]    -     -     -    -
   1    [1]   [2]   [5]   [4]  [0]
   2     -    [3]    -     -    -
   3    [1]   [3]   [5]   [8]   -
   4    [1]   [2]   [5]   [1]   -
   5     -     -    [6]    -    -
   6     -     -    [7]    -    -
   7    [1]   [2]   [7]   [4]   -
   8    [9]    -     -    [9]   -
   9    [1]   [2]   [5]   [9]   -
Nodes examined 28395, Nodes created 11052,
Completed PFSA 33
CPU (486-66) time: 0:00:49
```

Using this technique it is now not possible to prove a given solution is optimal, however in no observed case has a better solution been yielded without the compatibility testing. For most non-trivial problems it is not possible to exhaustively search the probable search space to obtain this proof anyway.

## 4.7 STOCHASTIC SEARCHING

The problem with selecting the node of lowest heuristic value is the tendency to become permanently trapped in local minima. A major reason for this permanent entrapment is that only a limited number of nodes in the construction tree can be maintained in memory, say 100,000-200,000, beyond this limit nodes have to be culled. This number of nodes represents only be a tiny fraction of the probable search space. If a local minima is reached, always selecting the node with the lowest heuristic value can quickly flood the pool of kept nodes.

To reduce the risk of becoming trapped in local minima and to help maintain diversity of the nodes stored in memory the probable search space is searched stochastically. There are two versions of stochastic sampling that are employed reasonably successfully. Only the *Tiered Probabilities* method is presented in this paper. The *Likelihood Probabilities* method is found in Collins and Oliver (1997b).

### 4.7.1 Tiered Probabilities

To select a node to expand the tree is traversed from the root node down to a leaf and that node is then expanded. A strategy for introducing randomness into the search of the probable search space is to select the best child node (lowest heuristic) with some probability ($\mu$) otherwise select a child at random. The value of $\mu$ is chosen at random before traversing the tree. There are 4 different values of $\mu$ that are selected. Table 6 gives the probability that a particular value of $\mu$ will be chosen. The values of $\mu$ and the probability of selection has been chosen on experimental basis.

Table 6: Distribution of selected values of μ.

| Value of μ | % of Searches Selected |
|---|---|
| 1.00 | 50 |
| 0.80 | 35 |
| 0.50 | 10 |
| 0.00 | 5 |

Table 7 gives the results obtained for the secondary structure data using the tiered probabilities. This is the best result that has been obtained. It is not know for this data set if a better PFSA exists.

Table 7: Secondary Structure Result: Switched heuristics, compatibility testing, tiered probabilities.

```
There are 12 states with a max of 5 arcs
Automata cost is: 12408.63456bits
arc->    O     E     H     T    d
 state
   0    [1]    -     -     -    -
   1    [1]   [2]   [5]   [4]  [0]
   2     -    [3]    -     -    -
   3    [1]   [3]   [5]   [10]  -
   4    [3]    -    [5]   [8]   -
   5     -     -    [6]    -    -
   6     -     -    [7]    -    -
   7    [1]   [2]   [7]   [9]   -
   8    [1]   [2]   [5]   [8]   -
   9    [1]   [2]   [5]   [8]   -
  10    [11]   -     -    [11]  -
  11    [1]   [2]   [5]   [11]  -
Nodes examined 270566, Nodes created
80865, Completed PFSA 1672
CPU (486-66) time: 0:09:26
```

## 5. COMPARISONS WITH OTHER MEASURES

The power of the IGS algorithm is readily put into context when it is compared to other finite state automata induction algorithms. The methods that the IGS algorithm is compared with are the *k*-tails of algorithm Biermann and Feldman (1972) and the *sk*-strings algorithm of Raman and Patrick (1995). The *sk*-strings algorithm of Raman and Patrick (1995) is assumed to be the current state of the art for the induction style of merging states based on equivalence relations. The NDFSA are determinised by the software provided with the packages.

The tests are performed by generating 25 random PFSA and then randomly sampling data from the generating PFSA until at least 4 transitions have occurred per arc. The random sample of data is then used to attempt to induce the original machine. The problems are intended to be difficult to analyse the performance of the algorithms for non-trivial problems. The performance of a particular induction is given by the MML ratio measure. The MML ratio measure is computed as the induced automaton's MML divided by the generating automaton's MML. Details of the random PFSA generating algorithm and justification for the MML ratio test are given in Collins and Oliver (1997b). An MML ratio of 1.0 or just under[7] indicates an exact match. An MML ratio of greater than 1.2 usually indicates a poor match. Table 8 shows the MML ratios achieved for each of the algorithms.

Table 8: Comparison of results.

| No. | States | Arcs | IGS | *sk*-strings | *k*-tails |
|---|---|---|---|---|---|
| 0 | 29 | 43 | 1.000 | 3.463 | 2.740 |
| 1 | 80 | 346 | 1.055 | DNF | DNF |
| 2 | 38 | 90 | 1.000 | DNF | DNF |
| 3 | 5 | 15 | 1.000 | 1.423 | 2.554* |
| 4 | 5 | 8 | 1.000 | 1.911* | 1.862* |
| 5 | 110 | 199 | 1.008 | 2.018 | 1.304 |
| 6 | 76 | 136 | 1.033 | 2.660 | 2.362 |
| 7 | 72 | 148 | 1.019 | DNF | DNF |
| 8 | 58 | 180 | 1.010 | DNF | DNF |
| 9 | 51 | 124 | 1.010 | DNF | DNF |
| 10 | 43 | 149 | 1.006 | DNF | 2.140* |
| 11 | 69 | 104 | 0.998 | 2.204 | 1.580 |
| 12 | 66 | 120 | 1.507 | DNF | DNF |
| 13 | 7 | 10 | 1.000 | 1.310 | 2.223* |
| 14 | 80 | 182 | 1.030 | DNF | DNF |
| 15 | 78 | 532 | 1.474 | DNF | DNF |
| 16 | 93 | 133 | 1.002 | DNF | DNF |
| 17 | 41 | 134 | 1.818 | DNF | DNF |
| 18 | 26 | 123 | 1.523 | DNF | DNF |
| 19 | 64 | 95 | 0.999 | DNF | DNF |
| 20 | 64 | 226 | 1.602 | DNF | DNF |
| 21 | 46 | 56 | 1.000 | 7.773 | 7.470 |
| 22 | 45 | 90 | 0.999 | 2.135 | 1.675 |
| 23 | 77 | 122 | 1.000 | DNF | DNF |
| 24 | 65 | 84 | 0.998 | DNF | DNF |

Notes:
*sk*-strings run with default parameters, *k*=1.
k-tails run with default parameters, k=3.
*: MML ratio worse than Null theory (1 state machine)
DNF: Program did not finish.
All programs executed on 200Mhz DEC Alpha.
Time out on IGS algorithm 1000 seconds. *Sk*-strings and *k*-tails 7 days.

From Table 8, the efficiency of the IGS algorithm can easily be observed when applied to non-trivial problems as compared to existing FSA induction algorithms. Of the 25 test runs the IGS algorithm produced 11 exact matches, 9 near matches and 5 failures (MML ratio > 1.2). From the 25 test runs, 16 were unable to complete for the *sk*-strings algorithm and 15 for the *k*-tails algorithm. By the MML ratio test of greater than 1.2, all solutions found by either algorithm would be considered failures. A number of the solutions found were worse than the null theory! Of the 9 trial automata the *sk*-strings algorithm did complete 7 exact matches were produces by the IGS algorithm and 2 very close matches.

7 An MML ratio of just under 1 occurs when the generating PFSA has a minor redundancy like two identical states and these are merged in the induced automaton.

Of the 14 trials that neither the *sk*-strings algorithm or the *k*-tails algorithm were able to complete 4 were exactly solved by the IGS algorithm and another 6 were fairly close.

For the *sk*-strings algorithm and the *k*-tails algorithm for small problems they would complete very rapidly (under 1 second), however once the sample data input size grew in size the canonical machine would often grow to thousands of states and the problem became unsolvable in the given time period.

# 6. EFFICIENCY AND DATA SET SIZE

A highly positive aspect of the IGS algorithm is that the efficiency of the algorithm improves as more data is provided and there is a better statistical representation of the machine to be learnt. As an example of the performance improvement we study a random machine that has 29 states with 7 token types. The internal structure of the machine consists of 117 arcs, 8 of which are delimited. The MML ratio is given in Table 9 for various data set sizes sampled from this machine.

Table 9: Effect of increased sample size.
Time out: 300s, 486-66.

| Sentences | Tokens | States | MML (bits) | MML Ratio |
|---|---|---|---|---|
| 183 | 2756 | 19 | 6816.7 | 1.222 |
| 264 | 3962 | 24 | 9655.7 | 1.245 |
| 301 | 4652 | 34 | 10955.6 | 1.218 |
| 444 | 6945 | 29 | 15601.1 | 1.196 |
| 799 | 12530 | 34 | 28157.6 | 1.219 |
| 1635 | 24898 | 29 | 45273.9 | 1.000 |
| 2989 | 45619 | 29 | 82521.2 | 1.000 |

For the given machine the algorithm is correctly able to identify an isomorphic equivalent (same machine, different state ordering) to the original machine in the 300 second time period with a sample size somewhere between 800 and 1635 sentences. For the sample size of 1635 the problem is solved with less than 1 second CPU time and only expanding 185 nodes. Increasing the sample size to 2989 sentences requires only 169 nodes to be expanded to find the correct solution.

# 7. CONCLUSION

In this paper we have introduced a new algorithm (IGS) for the inductive inference of PFSA from positive examples of behaviour. The IGS algorithm successfully bridges the gap between the current methodologies of automaton induction. These being: providing a criterion by which to choose between pairs of automaton and the provision of an algorithm to reduce the number of machines required to be examined. The IGS algorithm tightly couples these two areas by providing a criterion with strong mathematical background to drive an efficient search algorithm. To do this the algorithm takes advantage of the MML measure and applies it in a unique way, introducing the concept of partial MML to partition the model space.

When the algorithm is compared to some existing algorithms on difficult problems, the IGS algorithm has proved to be vastly superior in both run time and quality of induction.

## 7.1 FUTURE WORK

From our studies we are encouraged by the belief that the IGS algorithm has much scope for further improvement. The efficiency of the algorithm hinges on four main areas and these are considered individually and are examined in order of perceived importance.

1) MML Estimation: The MML estimation technique forms the basis for an efficient search heuristic. The MML estimation technique presented in this paper is fairly straight forwards. It would seem promising to apply a technique that takes into account more localised generation tree information and likely error of the MML estimate.

2) Culling Heuristics: The compatibility testing heuristic proved to be a very powerful technique for assisting the search algorithm. The problem with this technique is that it only looks at the distribution of next actions on dangling arcs. There can be at most $S$ (where $S$ is the number of sentences in $D$) transitions on any one of these arcs and usually considerably less. With a small sample size it is difficult to say that two distributions are dissimilar with a high degree of certainty. It is proposed that a possible addition to the compatibility testing heuristic is to periodically recheck complete arcs when many more transitions are available.

3) Minima Avoidance: The algorithm becomes trapped in local minima when too many nodes in memory are from the same section of the generation tree. Better techniques are required to maintain diversity in the memory store. This may require sacrificing nodes with a lower estimated MML to nodes that are from different parts of the generation tree.

4) Criterion: If two criteria generally agree as to which PFSA is the best then what distinguishes criteria is how well they spread the range of MML values. Using a criterion with a better spread results in less penalty being paid by errors in the MML estimation search heuristic.

## Acknowledgments

We would like to thank Anand Raman[8] for providing public access to the **sk-strings**, **k-tails** and **dfa** software available from the web page.

http://fims-www.massey.ac.nz/~ARaman/

[8] Department of Computer Science, Massey University, Palmerston North Campus, New Zealand. A.Raman@massey.ac.nz